\documentclass{bmvc2k}

\usepackage{amsmath,amssymb}
\usepackage{amsfonts}
\usepackage{makecell}
\usepackage{multirow}
\usepackage{subfig}
\usepackage{bm}
\usepackage{booktabs}
\usepackage{standalone}
\usepackage{algorithm}
\usepackage{algpseudocode}
\usepackage{courier}
\usepackage{verbatim}
\usepackage[table]{xcolor}

\title{Residual Likelihood Forests}

\addauthor{Yan Zuo}{yan.zuo@monash.edu}{1}
\addauthor{Tom Drummond}{tom.drummond@monash.edu}{1}

\addinstitution{
ARC Centre of Excellence for Robotic Vision, Monash University, Australia
}

\runninghead{Yan Zuo, Tom Drummond}{Residual Likelihood Forests}

\begin{document}
\maketitle
\begin{abstract}
This paper presents a novel ensemble learning approach called Residual Likelihood Forests (RLF). Our weak learners produce conditional likelihoods that are sequentially optimized using global loss in the context of previous learners within a boosting-like framework (rather than probability distributions that are measured from observed data) and are combined multiplicatively (rather than additively). This increases the efficiency of our strong classifier, allowing for the design of classifiers which are more compact in terms of model capacity. We apply our method to several machine learning classification tasks, showing significant improvements in performance. When compared against several ensemble approaches including Random Forests and Gradient Boosted Trees, RLFs offer a significant improvement in performance whilst concurrently reducing the required model size.
\end{abstract}

\section{Introduction}
\label{sec:intro}
Ensemble and Boosting methods such as Random Forests~\cite{breiman2001random} and AdaBoost~\cite{schapire1999brief} are often recognized as some of the best out-of-the-box classifiers, consistently achieving state-of-the-art performance across a wide range of computer vision tasks including applications in image classification~\cite{bosch2007image}, semantic segmentation~\cite{shotton2008semantic}, object recognition~\cite{gall2013class} and data clustering~\cite{moosmann2006fast}. The success of these methods is attributed to their ability to learn models (strong learners) which possess low bias and variance through the combination of weakly correlated learners (weak learners). Forests reduce variance through averaging its weak learners over the ensemble. Boosting, on the other hand, looks towards reducing both bias and variance through sequentially optimizing under conditional constraints. 

The commonality between both approaches is in the way each learner is constructed: both methods use a top-down induction algorithm (such as CART~\cite{breiman1984classification}) which greedily learns decision nodes in a recursive manner. This approach is known to be suboptimal in terms of objective maximization as there are no guarantees that a global loss is being minimized~\cite{hastie2009elements}. In practice, this type of optimization requires the non-linearity offered by several (very) deep trees, which results in redundancy in learned models with large overlaps of information between weak learners. 



To address these limitations, the ensemble approaches of~\cite{friedman2001greedy,schulter2013alternating} have utilized gradient information within a boosting framework. This allows weak learners to be fit via pseudo-residuals or to a set of adaptive weights and allows for the minimization of a global loss via gradient descent. Whilst these methods have demonstrated improved performance over their Random Forest counterpart, they are more susceptible to overfitting, making them harder to tune. An alternative approach proposed by~\cite{ren2015global} tried to retain Random Forests as a baseline classifier, employing a global refinement technique combined with leaf pruning to gain improvements in performance and reduce model size. However, this method still required the relatively large overhead of constructing the original Random Forest before refinement can be performed (with an additional overhead associated with global refinement and pruning).

In this work,  we propose a method which constructs a decision forest that benefits from the boosting approach of creating complementary weak learners through minimizing a global loss, yet retains the simplicity in model tuning of Random Forests. Our approach monotonically decreases the loss function of the ensemble using a novel, top-down induction algorithm, which unlike CART, accounts for the mutual information between weak learners already in the ensemble. Empirically, we extensively evaluate our approach across multiple datasets against several ensemble and boosting baselines. Our results show that our method improves classification performance at a higher parameter efficiency.

The main contributions of this paper are as follows:
\begin{itemize}
\item We propose a novel ensemble learning framework that includes a novel induction algorithm which accounts for the mutual information between weak learners it constructs.
\item Theoretically, we show that our method allows optimization to occur in closed-form (in contrast to existing gradient boosting methods), which simplifies model tuning.
\item Empirically, we show that our method offers dramatic reductions in required model size. Further, we show that our approach outperforms several competing boosting and ensemble baselines across 5 standard machine learning datasets. 
\end{itemize}
\section{Related Work}
\label{sec:related_work}
\subsection{Random Forests}
\label{ssec:random_forests}
RFs became popular in the literature due to their flexibility and robustness when dealing with data-driven learning problems~\cite{caruana2006an}. RFs were first developed by~\cite{ho1995random} to overcome the variance and stability issues found in binary decision trees by randomly choosing the subspace of features for tree training. This method was refined by~\cite{breiman2001random}, where the bootstrap aggregating technique of~\cite{breiman1996bagging} was combined with the random feature selection of RFs.~\cite{geurts2006extremely} developed Extremely Randomized Trees, where both the attribute and cut-point of the input feature were randomized.~\cite{shotton2008semantic} applied RFs to the semantic segmentation problem, using them as a global image classifier to quickly and efficiently provide priors to be pipelined into the system developed in~\cite{shotton2006textonboost}.

\subsection{Boosting}
\label{ssec:boosting}
The work of~\cite{schapire1990the} first introduced the concept of boosting using an algorithm that selectively combined several weak hypotheses into a single strong hypothesis.~\cite{freund1990boosting} extended this algorithm to combine the outputs of all weak hypotheses. These methods however, did not take advantage of the adaptability of weak learners. This was addressed by~\cite{freund1995a} with Adaptive Boosting, where each subsequent weak learner introduced to the strong learner is biased to favor misclassified training samples using high-weight data points.~\cite{breiman1998arcing} extended on this concept of adaptable boosting by offering an alternative to using weights, formulating the method as a gradient descent problem with a specialized loss function. 
\section{Preliminaries}
\label{sec:preliminaries}

Conventional decision trees consist of internal decision nodes and terminal leaf nodes. The internal nodes of each decision tree are a set of $\mathcal{N}_d$ decision nodes, $D=\{d_0, \cdots, d_{\mathcal{N}_d-1}\}$, each of which holds a decision function $d_{i}(\bm{x}; \theta_{i})$, where $\theta_{i}$ are the parameters for decision node $i$ in the tree. For binary decision trees, the decision function for each decision node $i$ is defined as $d_{i}(\bm{x}; \theta_{i}) : \mathcal{X} \rightarrow [0, 1]$, which routes a sample to a left or right subtree. Collectively, the decision nodes route data, $\bm{x}$, through the tree until a terminal leaf node is reached: $\ell = \delta(\bm{x};\Theta)$. Here, $\delta$ signifies the routing function which directs data $x$ to the terminal leaf node $\ell$, given the tree's decision node parameters $\Theta$. 

For classification, the leaf nodes $Q$ in a decision tree contain class label probability distributions, $\bm{q}=Q(\ell)$. These distributions are formed from the training data and accompanying ground truth class labels. For a terminal leaf node $\ell$, the stored probability value for class $j$ is given as:
\begin{equation}
\label{class_label_dist}
q_j = Q(\ell)_j = \frac{\sum_n^N c_{n,j}}{N_\ell}
\end{equation}
where $N_{\ell}$ is the total number of samples routed into leaf node $\ell$ and $c_{n,j}$ is the observed class label for sample $n$ in class $j$:
\begin{equation}
\label{eq:class_label_def}
    c_{n,j} = 
    \begin{cases}
        1,              & \text{if sample $n$ has class label $j$} \\
        0,              & \text{otherwise}
    \end{cases}
\end{equation}

Typically, the decision nodes of the tree are chosen such that resulting distributions generated give the best split according to some purity measure (\textit{i.e.} entropy or the Gini index~\cite{breiman2001random}). 

A conventional decision forest is an ensemble of $\mathcal{N}_t$ number of decision trees. It delivers a final prediction for the input $\bm{x}$ by routing $\bm{x}$ through each decision tree in the ensemble to a leaf node and taking the average prediction delivered by each tree:
\begin{equation} 
\label{eq:forest_output}
P_j = \frac{1}{\mathcal{N}_t}\sum_{t}^{\mathcal{N}_t}
Q^t(\delta^t(\bm{x};\Theta^t))_j
\end{equation}
where $Q^t$, $\delta^t$ and $\Theta^t$ are the respective distributions, routing functions and parameters of tree $t$ in the ensemble.

\section{Residual Likelihood Forests}
\label{sec:residual_likelihood_forests}

First, we motivate our proposed approach using the following simple classification example involving two weak learners evaluating some input data. To keep our notation concise, we denote ${}^1\! p_j$ and ${}^2\! p_j$ to represent the probabilities assigned by each weak learner for the underlying class label $j$ given the input. Given that the prior is uniform over classes, the correct approach to combining these two weak learners depends on the correlation between the assigned probabilities from each weak learner. 

If ${}^1\! p_j$ and ${}^2\! p_j$ are independent of one another, the correct class probability should be given by the normalized product for the class $j$:
\begin{equation}
\label{eq:multiplicative_distributions}
P_j = \frac{{}^1\! p_j {}^{2}\!p_j}{\sum_k^{\mathcal{N}_c} {}^1 \!p_k {}^2\! p_k}
\end{equation}
where $\mathcal{N}_c$ is the number of classes. However, when the two weak learners are fully correlated, the assigned probabilities from each weak learner would be the same and ${}^1\! p_j = {}^2\! p_j$. In this instance, applying Eq.~\ref{eq:multiplicative_distributions} would result in an incorrect estimate of the class probabilities as the normalized squared distribution and we should instead simply average their marginal distributions. 

\subsection{Weak Learners Generating Likelihoods}
\label{ssec:residual_likelihoods}
In practice, probability distributions $\bm{}^1\!\bm{p}$ and ${}^2\!\bm{p}$ will rarely be fully independent or fully correlated. As such, deciding the amount each weak learner should contribute to the strong classifier's prediction is a non-trivial task. Typically, ensemble methods combine their weak learners by averaging their marginal distributions~\cite{schapire1990the,freund1990boosting,breiman1996bagging}, relying on the law of large numbers to get a good estimate of the underlying distribution of the data and smooth out variance between learners.

The correlation between learners can be problematic if we wish to have a consistent approach for correctly combining weak learners within an ensemble since the approach that should be adopted depends on this correlation. Since it is difficult to determine whether weak learners are learning independent information or correlated information, we approach the problem from a different perspective: instead of deciding how to combine distributions from each weak learner, we can instead modify the stored $p$ distributions of each weak learner in a manner which allows for a consistent approach towards combining each weak learner's contribution irrespective of their correlation. 

Looking back at the simple classification example, if ${}^1\! p_j = {}^2\! p_j$, modifying the stored distributions $p$ into information vectors $\mathcal{q}$ by replacing ${}^1\! p_j$ and ${}^2\! p_j$ with their square roots and applying Eq.~\ref{eq:multiplicative_distributions} gives the correct answer for ${}^1\! p_j = {}^2\! p_j$:
\begin{equation}
P_j = \frac{{}^1\! \mathcal{q}_j {}^{2}\!\mathcal{q}_j}{\sum_k {}^1 \!\mathcal{q}_k {}^2\! \mathcal{q}_k} = \frac{\sqrt{{}^1\! p_j} \sqrt{{}^{2}\!p_j}}{\sum_k \sqrt{{}^1 \!p_k} \sqrt{{}^2\! p_k}}
\\ = {}^1 \!p_j= {}^2 \!p_j
\label{eq:sq_root_distributions}
\end{equation}

These information vectors ${}^1\! \bm{\mathcal{q}}$ and ${}^{2}\! \bm{\mathcal{q}}$ no longer represent probability distributions. Instead, each information vector can be learned and combined with other information vectors contributed by other weak learners from the ensemble. This forms the basis for our ensemble learning framework: for each decision tree, we construct the information vectors $\{\bm{\mathcal{q}}_1, ..., \bm{\mathcal{q}}_{\mathcal{N}_l}$\} for its $\mathcal{N}_l$ leaf nodes. Each vector is conditioned upon the prior information contained by trees already existing in the ensemble and is constructed to contain any new residual information gained from adding the new tree. and henceforth, we refer to each vector $\bm{\mathcal{q}}$ as a \textit{residual likelihood}. As we will demonstrate shortly, we can generate these residual likelihoods such that they can be treated as independent sources of information, forming a strong classifier distribution by taking their product and normalizing:
\begin{equation} 
\label{eq:multiplicative_forest}
P_j = \frac
{\prod_{t}^{\mathcal{N}_t}\mathcal{q}_{t,j}}
{\sum_k \prod_{t}^{\mathcal{N}_t} \mathcal{q}_{t,k}}
\end{equation}

\subsection{Optimization Method}
\label{ssec:residual_forest_framework}
To construct our forest, we minimize a global loss function in a similar manner to gradient boosting approaches~\cite{friedman2001elements}. The key difference in our approach is that the importance allocated towards hard-to-classify samples during the training process is implicit rather than explicit (as in the case of gradient boosted trees) and is directly incorporated into our induction algorithm. Specifically, our method still chooses decisions from a random subset of the feature space but these decisions are selected with the objective of minimizing the global loss function. Incidentally, this removes the need for weighted training samples or specialized loss functions found in traditional boosting techniques and simplifies the model tuning process.

\paragraph{Deriving the Objective Loss}
For a new tree to be added to the ensemble, we can derive a solution which optimizes the residual likelihood its leaf node contributes. Consider the $i^{th}$ tree to be added to the ensemble. We wish to find the residual likelihood for class $j$ ($\mathcal{q}_{i,j}$) it contributes such that the objective loss is minimized when this tree is added to the ensemble. For a given leaf node, we define the following terms:
\begin{itemize}
    \item $P_{n,j}^-$ is the prior value of class $j$ for sample $n$, combining likelihoods \{$\mathcal{q}_{1,j},...,\mathcal{q}_{i-1,j}$\} from all existing weak learners in the ensemble (excluding the new tree):
    \begin{equation}
    \label{combined_residual_likelihoods}
    P_{n,j}^- = \prod_{t}^{i-1}\mathcal{q}_{t,j}
    \end{equation}
    \item $\mathcal{q}_{i,j}$ is the stored value for class $j$ of the residual likelihood to be contributed by the new tree $i$.
    \item $P_{n,j}^+$ is the posterior probability of class $j$ for sample $n$ after new tree $i$ has contributed to the ensemble:
    \begin{equation}
    \label{averaged_prob_dist}
    P_{n,j}^+ = \frac{P_{n,j}^- \mathcal{q}_{i,j}}{\sum_k P_{n,k}^- \mathcal{q}_{i,k}}
    \end{equation}
    \item $c_{n,j}$ is the ground truth class label for sample $n$ taking the class label $j$, as defined in Eq.~\ref{eq:class_label_def}.
\end{itemize}

The cross entropy loss of the strong classifier after adding the new tree $i$ is defined as:
\begin{equation}
\label{eq:loss_function_1}
\mathbb{L} = -\sum_{n}\sum_{j} c_{n,j} \log{\left(P_{n,j}^+\right)}
\end{equation}
After applying Eq.~\ref{averaged_prob_dist}, we obtain the following:
\begin{equation}
\label{eq:loss_function_2}
\begin{split}
\mathbb{L} &= -\sum_{n}\sum_{j} c_{n,j} \log{\left(\frac{P_{n,j}^- \mathcal{q}_{i,j}}{\sum_k P_{n,k}^- \mathcal{q}_{i,k}}\right)} \\
&= -\sum_{n}\sum_{j} c_{n,j} \left(\log{\left(P_{n,j}^- \mathcal{q}_{i,j}\right)} - \log{\left(\sum_k P_{n,k}^- \mathcal{q}_{i,k}\right)}\right) 
\end{split}
\end{equation}
For the $i^{th}$ tree added to the strong classifier, we require a contributed residual likelihood $\mathcal{q}_{i,j}$ to the class $j$ such that $\mathbb{L}$ is minimized. For the class $j$, the first-order derivatives of Eq.~\ref{eq:loss_function_2} is given as:
\begin{equation}
\label{eq:first_order_derivative}
\frac{\partial \mathbb{L}}{\partial \mathcal{q}_{i,j}} = -\sum_{n} c_{n,j} \frac{P_{n,j}^-}{P_{n,j}^- \mathcal{q}_{i,j}} + \sum_{n} \frac{P_{n,j}^-}{\sum_{k} P_{n,k}^- \mathcal{q}_{i,k}}
\end{equation}
Where we set Eq.~\ref{eq:first_order_derivative} to zero to yield a solution:
\begin{equation}
\label{eq:residual_solution}
\begin{split}
\frac{N_j}{\mathcal{q}_{i,j}} &= \sum_{n} \frac{P_{n,j}^-}{\sum_{k} P_{n,k}^- \mathcal{q}_{i,k}} \\
\therefore N_j &= \sum_{n} \frac{P_{n,j}^- \mathcal{q}_{i,j}}{\sum_{k} P_{ik}^- \mathcal{q}_{i,k}} \\
\therefore N_j &= \sum_{n} P_{n,j}^+
\end{split}
\end{equation}
Here, $N_j$ is the total number of samples with class label $j$ within the given leaf node. This is a convex optimization problem with a single global solution where the loss function is convex:
\begin{equation}
\label{eq:convex_function}
\begin{split}
\frac{\partial^{2} \mathbb{L}}{\partial \mathcal{q}_{i,j}^{2}} = \sum_{n} & c_{n,j} \frac{1}{\mathcal{q}_{i,j}^{2}} - \sum_{n} \frac{(P_{n,j}^-)^{2}}{(\sum_{k} P_{n,k}^- \mathcal{q}_{i,k})^{2}} \geq 0 \\
\therefore \sum_{n} & c_{n,j} \geq \sum_{n} \frac{(P_{n,j}^-\mathcal{q}_{i,j})^{2}}{(\sum_{k} P_{n,k}^- \mathcal{q}_{i,k})^{2}} = \sum_{n} (P_{n,j}^+)^{2}
\end{split}
\end{equation}
Eq.~\ref{eq:convex_function} holds true for each sample $n$ where $0 < \sum_{k} (P_{n,k}^+)^{2}\leq1$ and $\sum_{k} c_{n,k}=1$, which satisfies the criteria for convexity.

\begin{algorithm}[t]
\caption{RLF Optimization}
\label{alg:learning_method}
\begin{algorithmic}[1]
\Require $\mathcal{N}$: training set
\Require Number of trees in forest $\mathcal{N}_t$
\Require Feature pool size $S$
\Require Maximum tree depth $D_{max}$
\ForAll {$t \in \{1,...,\mathcal{N}_t\}$}
    \ForAll {$m \in \{1,...,D_{max}\}$}
        \State In parallel:
        \ForAll {$s \in \{1,...,2^{D_{max}-1}\}$}
            \State In parallel:
            \ForAll {$p \in \{1,...,S\}$}
                \State Choose a random $\theta_p$
                \State Calculate $N_j$ and $\sum_i P_{n,j}^-$
                \State Calculate leaf residuals $\mathcal{q}_j$
                \State Calculate $\mathbb{L}$
            \EndFor
            \State Choose $\theta_p$ for node $d$ that minimizes $\mathbb{L}$
        \EndFor
    \EndFor
    \State Copy the winning $\mathcal{q}_j$ into the leaves of the tree $t$
\EndFor
\end{algorithmic}
\end{algorithm}

\paragraph{Computing Residual Likelihoods}
Eq.~\ref{eq:residual_solution} indicates that for a leaf node and samples routed to that leaf node, the solution for the residual likelihood $\bm{\mathcal{q}}$ exists when the sum of strong classifier probabilities across samples is equal to the sum of class labels associated with those samples for a given class. For computing a $\bm{\mathcal{q}}$ that fulfills this criteria, we adopt a straightforward strategy: using Eq.~\ref{eq:residual_solution}, we first initialize the residual likelihoods of the new tree $\bm{\mathcal{q}} = \bm{1}$ and solve for the updated residual:
\begin{align}
\label{eq:update_residual}
\text{Initialize: } & \mathcal{q}_j \leftarrow 1 \\
\label{eq:update_ql}
\text{Iterate: } & \mathcal{q}_j \leftarrow \mathcal{q}_j \frac{N_j}{\sum_{n} P_{n,j}^+}
\end{align}
which converges quickly and stabilizes when Eq.~\ref{eq:residual_solution} holds, solving for $\mathcal{q}_j$ for all $j$. In theory, the estimate for $\mathcal{q}_j$ at the solution can be improved arbitrarily by continuously iterating on Eq.~\ref{eq:update_ql}. However, in practice, we found that only a single iteration was required for Eq.~\ref{eq:residual_solution} to hold. In Fig~\ref{alg:learning_method}, we outline our approach.

\section{Experiments}
\label{sec:experiments}

\paragraph{Experiment Settings}
For all our experiments, we follow the settings described in~\cite{schulter2013alternating}. We set the default number of trees in an ensemble to 100, keeping in line with the experiments in~\cite{schulter2013alternating,ren2015global}. We use the XGBoost~\cite{chen2016xgboost} implementation of Gradient Boosted Trees (GBT), using the prescribed default settings. For completeness, we also trained an additional GBT model (XGBoost-Large) in which we match the number of trees and tree depth of our RLF model. 

For each dataset, we test a number of random features equal to the square root of the feature dimensions, as recommended in~\cite{breiman2001random}, allocating 10 random thresholds per feature. In each case, we report mean error and standard deviation across 10 separate runs to account for variance due to randomness during training, except for the \textit{G50c} dataset which we report the mean error and standard deviation across 250 separate runs as was done in~\cite{schulter2013alternating}. 

Due to our training scheme differing from the conventional entropy minimization approaches of competing approaches, we do not specify an early node termination policy. Instead, we rely on the efficiency of our decision node split choices and construct considerably shallower models. Hence, for practical purposes, we construct trees up to a maximum specified depth of 15. For the overall results shown in Table~\ref{tab:overall_comparison}, we follow the tree depth settings in~\cite{schulter2013alternating}, except for the \textit{Chars74k} dataset in which we set a tree depth of 15 (as opposed to 25).

\begin{figure*}[h]
\begin{center}
 \subfloat[]{%
  \begin{tabular}{c}
  \includegraphics[width=0.3\textwidth]{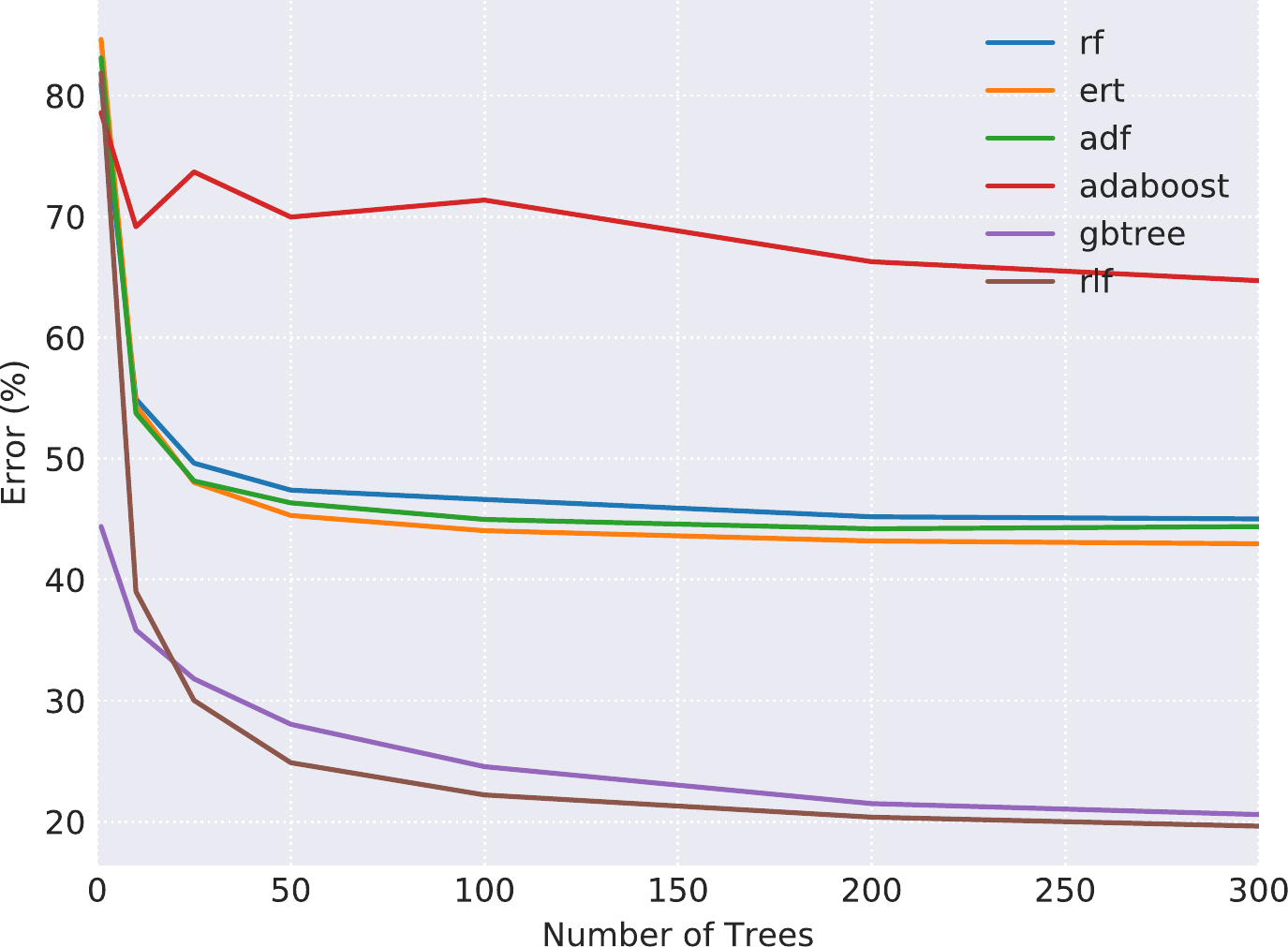}
  \end{tabular}
  \label{fig:char74k_num_trees_5depth_error}
 }%
 \hskip -2ex
 \subfloat[]{%
  \begin{tabular}{c}
  \includegraphics[width=0.3\textwidth]{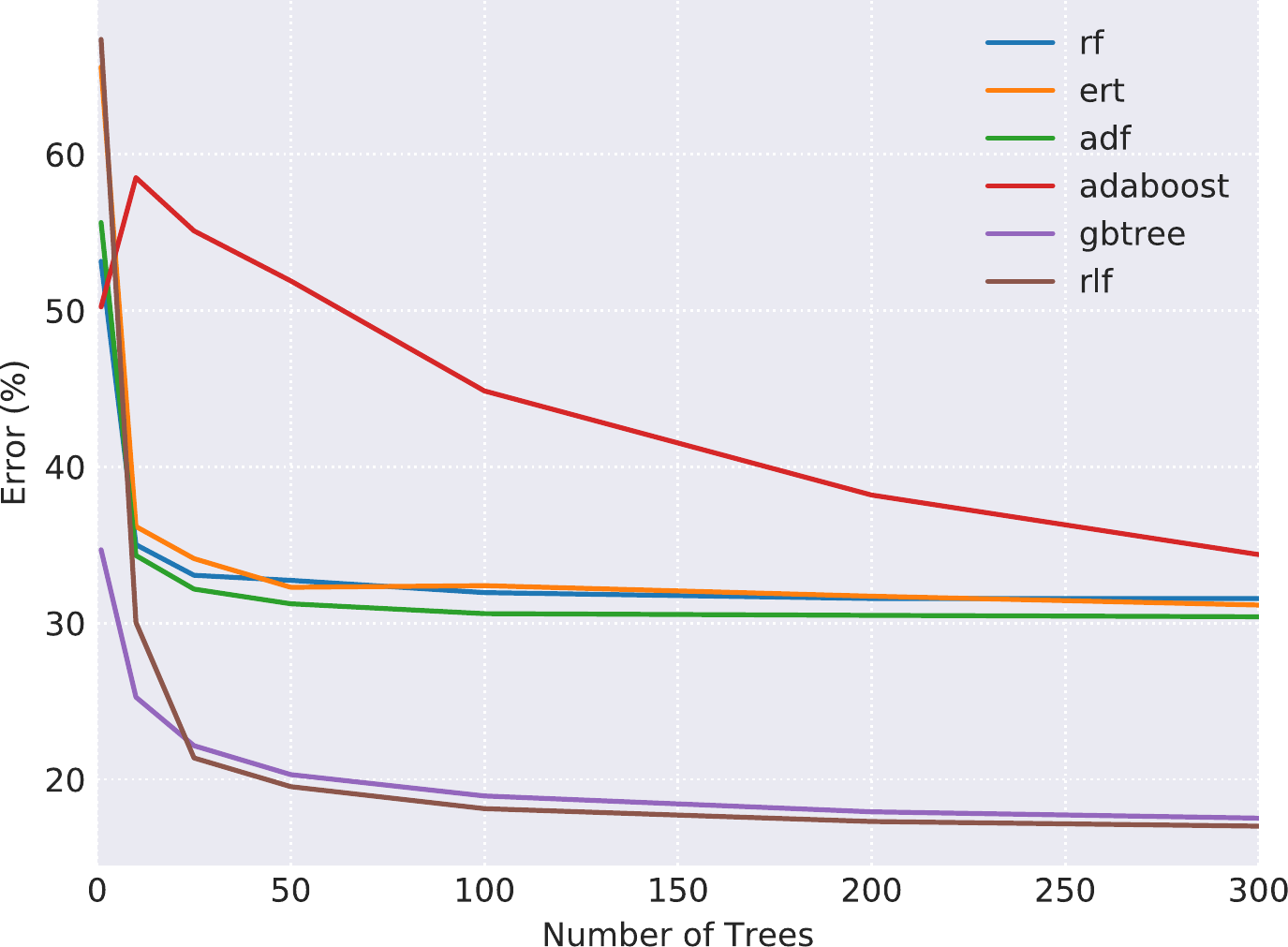}
  \end{tabular}
  \label{fig:char74k_num_trees_10depth_error}
 }%
 \hskip -2ex
 \subfloat[]{%
  \begin{tabular}{c}
  \includegraphics[width=0.3\textwidth]{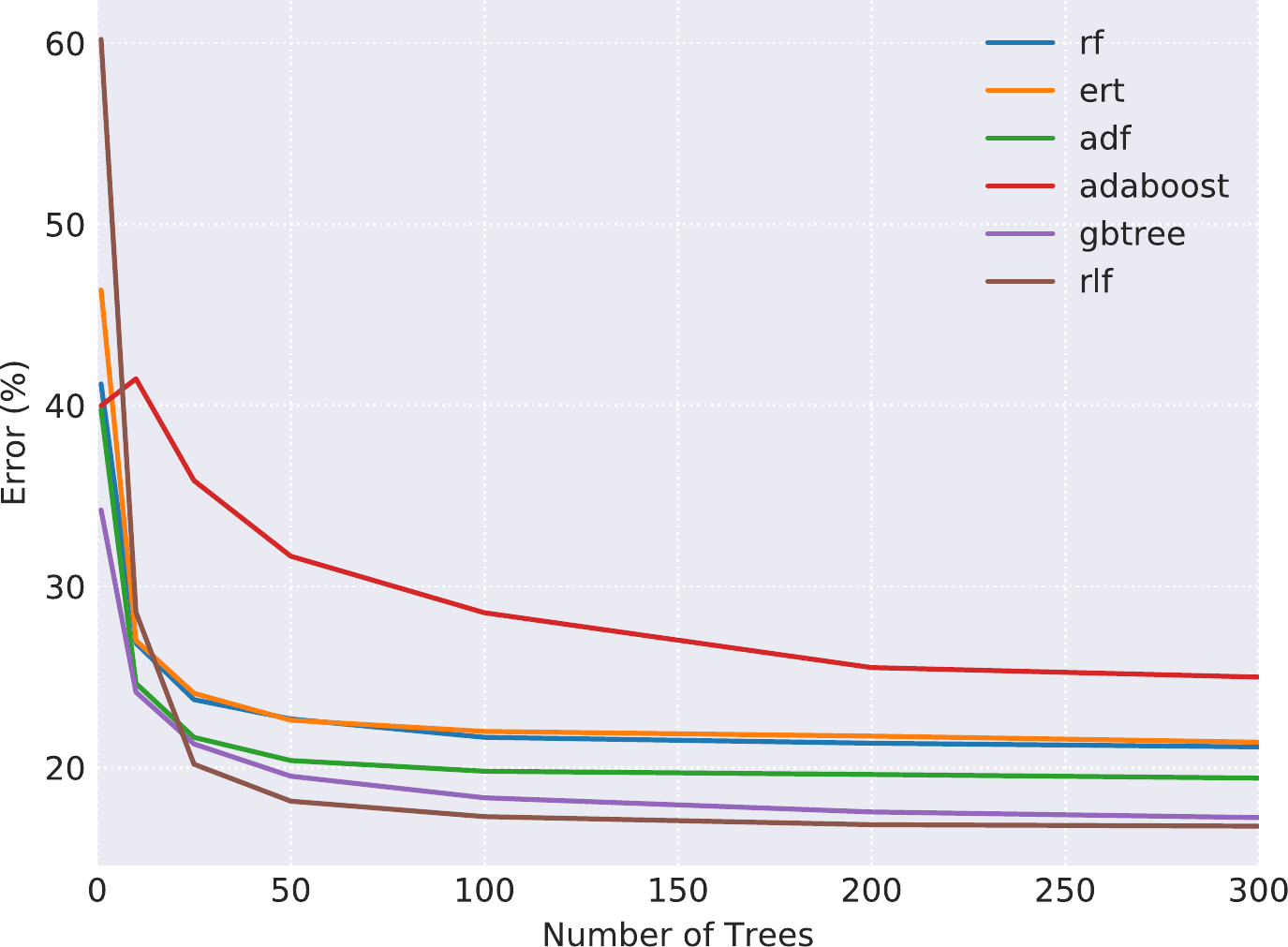}
  \end{tabular}
  \label{fig:char74k_num_trees_15depth_error}
 }%
 \vspace{1.0em}
\caption{Classification error of RLF versus competing baselines, where the number of trees is varied from $[1,300]$ for a fixed maximum tree depth of (a) 5, (b) 10 and (c) 15 on the \textit{Chars74k} test data.}
\label{fig:char74k_vary_depth}
\end{center}
\end{figure*}

\subsection{Chars74k \& MNIST Ablations}

\paragraph{Chars74k}
We vary hyperparameters of RLFs and competing baselines for the \textit{Chars74k} dataset, observing the classification error on the test data. The results of these experiments are shown in Fig.~\ref{fig:char74k_vary_depth}. Figs.~\ref{fig:char74k_num_trees_5depth_error},~\ref{fig:char74k_num_trees_10depth_error} and~\ref{fig:char74k_num_trees_15depth_error} show the test classification error on the \textit{Chars74k} dataset when maximum tree depth is restricted to $5$, $10$ and $15$ respectively. We can see that our RLF approach clearly outperforms all competing baselines, particularly with shallow trees. Only Gradient Boosted Trees (GBT) offers a competitive result, although we note that our method far easier to train than GBT, which we found was extremely sensitive to hyperparameters such as the learning rate.

\begin{figure*}[t]
\begin{center}
 \subfloat[]{%
  \begin{tabular}{c}
  \includegraphics[width=0.3\textwidth]{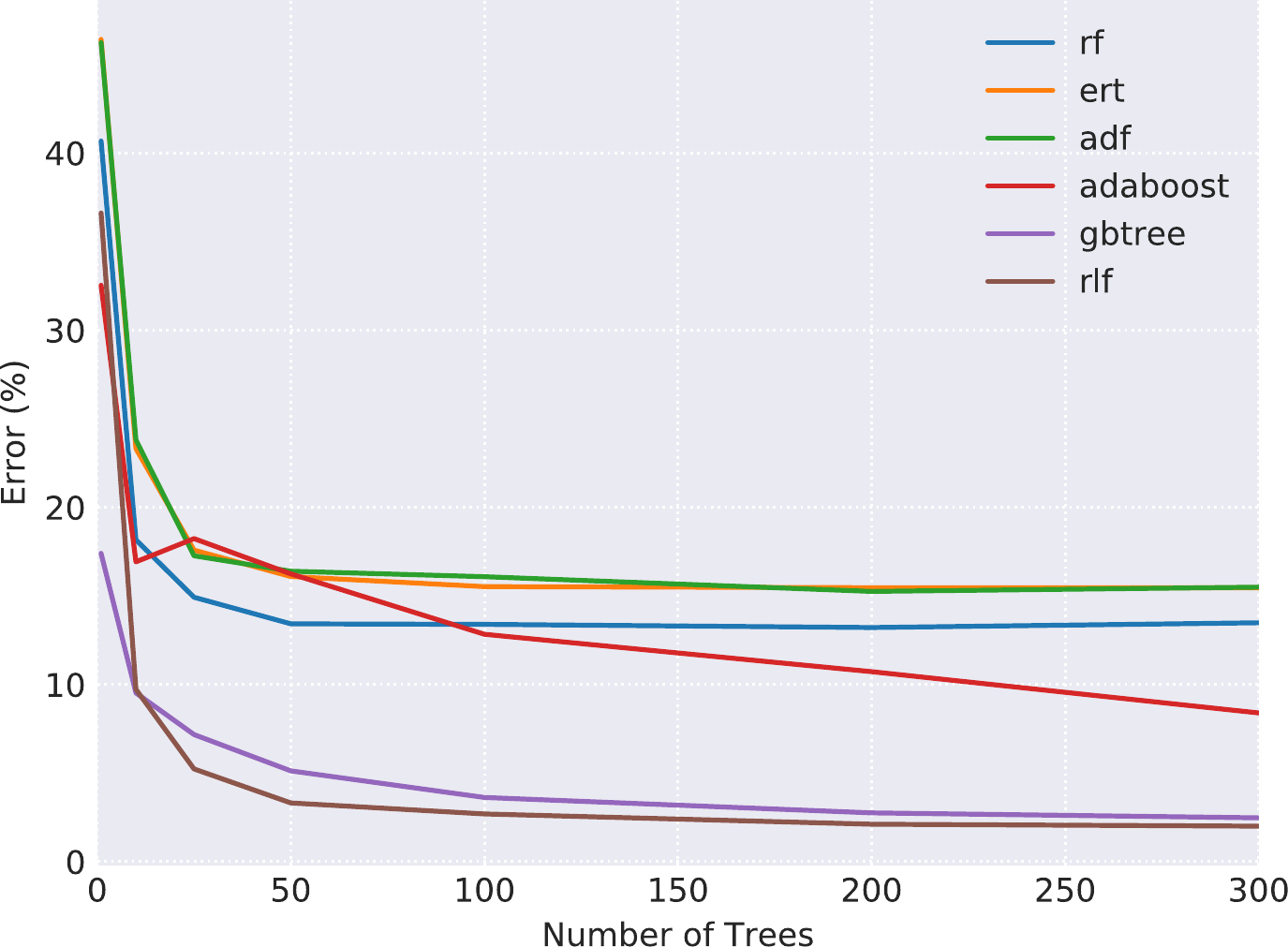}
  \end{tabular}
  \label{fig:mnist_num_trees_5depth_error}
 }%
 \hskip -2ex
 \subfloat[]{%
  \begin{tabular}{c}
  \includegraphics[width=0.3\textwidth]{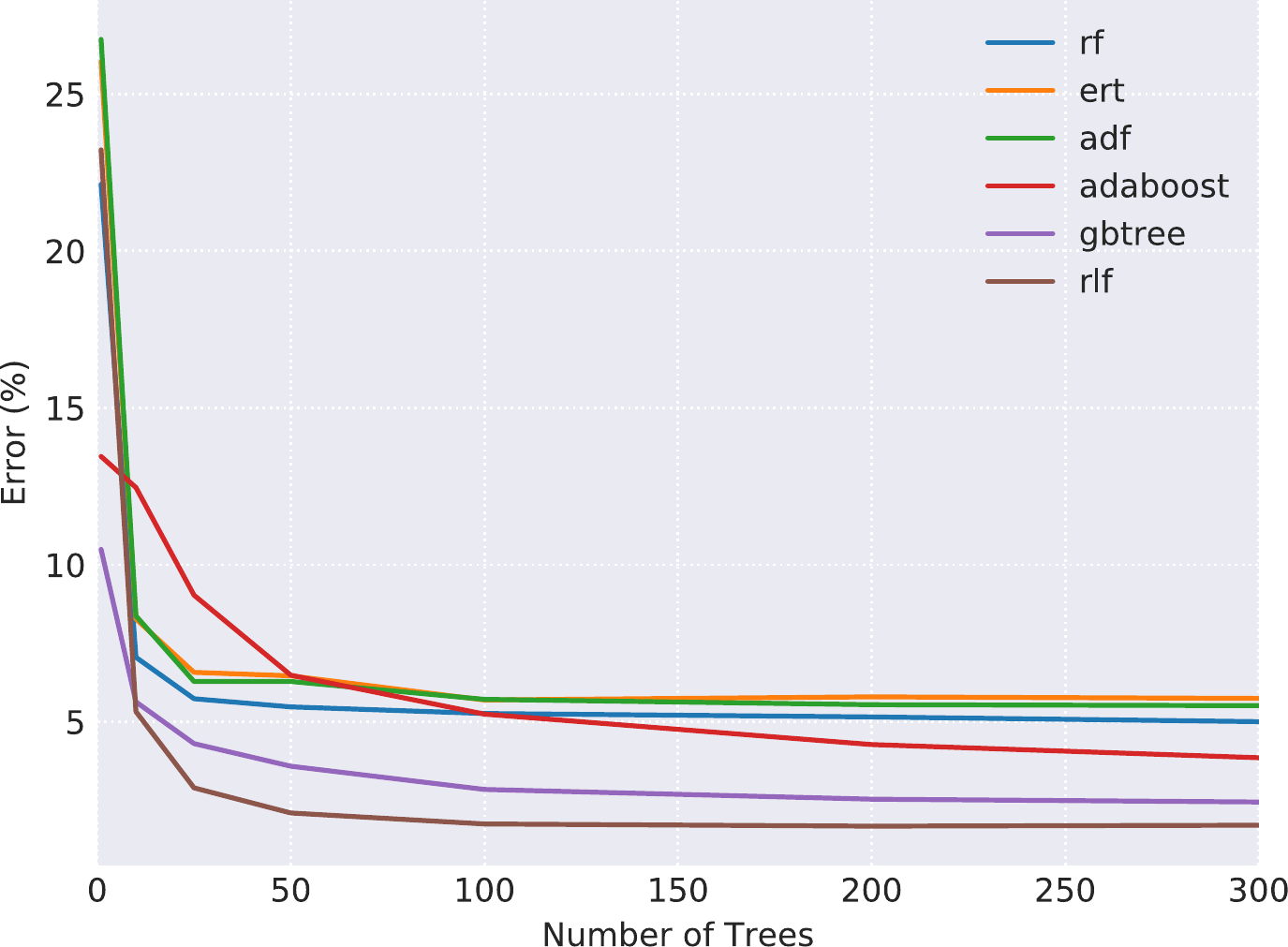}
  \end{tabular}
  \label{fig:mnist_num_trees_10depth_error}
 }%
 \hskip -2ex
 \subfloat[]{%
  \begin{tabular}{c}
  \includegraphics[width=0.31\textwidth]{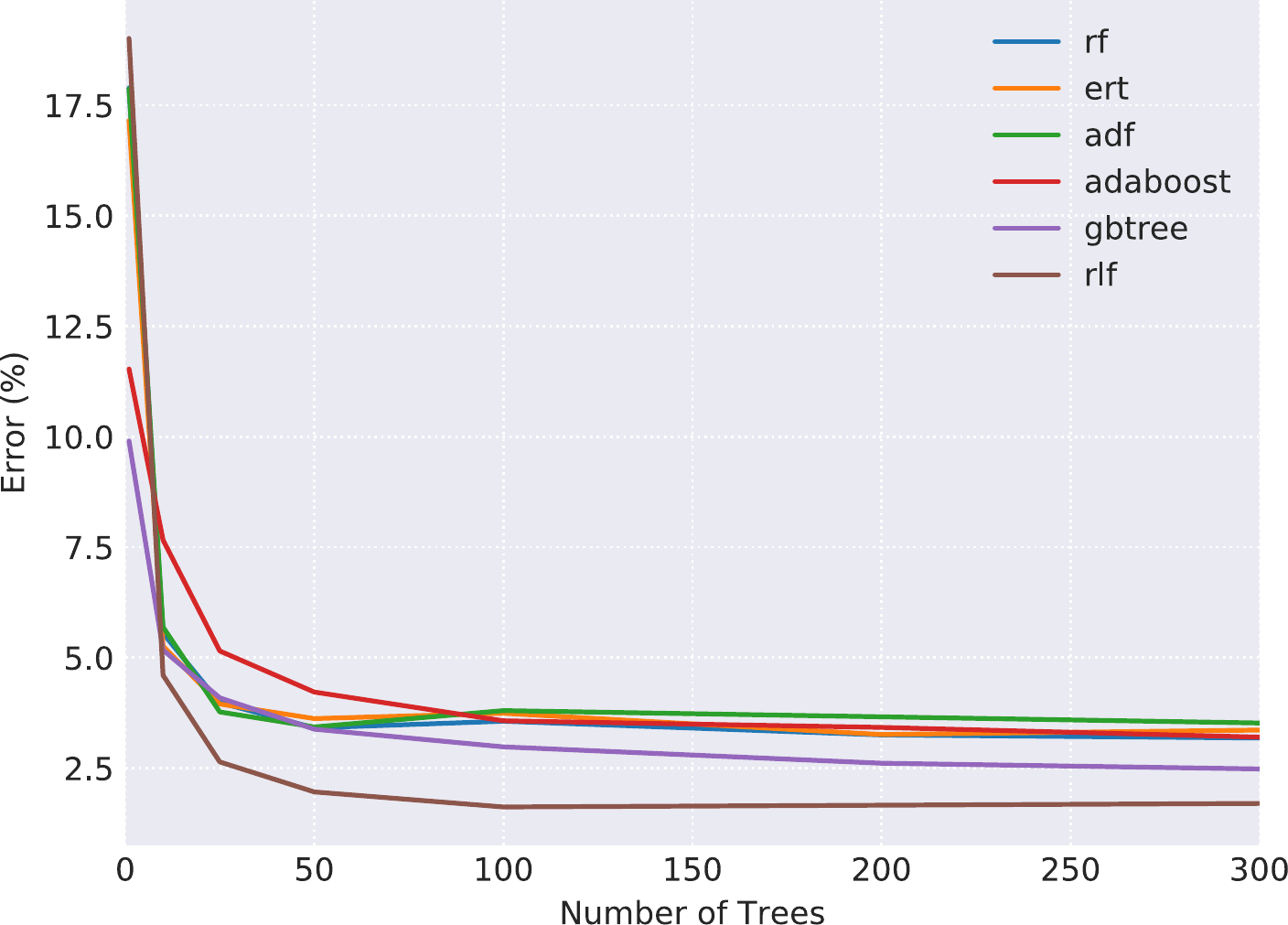}
  \end{tabular}
  \label{fig:mnist_num_trees_15depth_error}
 }%
 \vspace{1.0em}
\caption{Classification error of RLF versus competing baselines, where the number of trees is varied from $[1,300]$ for a fixed maximum tree depth of (a) 5, (b) 10 and (c) 15 on the \textit{MNIST} test data.}
\label{fig:mnist_vary_depth}
\end{center}
\end{figure*}

\paragraph{MNIST}
Furthermore, we compare RLFs to competing baselines for the \textit{MNIST} dataset and observe classification error on the test data. These ablations are shown in Fig.~\ref{fig:mnist_vary_depth}. In Figs.~\ref{fig:mnist_num_trees_10depth_error},~\ref{fig:mnist_num_trees_10depth_error} and~\ref{fig:mnist_num_trees_15depth_error}, we show the test classification error on the \textit{MNIST} dataset when maximum tree depth is restricted to $5$, $10$ and $15$ respectively. We can see again our RLF approach clearly outperforms all competing baselines, particularly with shallow trees. Similarly, Gradient Boosted Trees (GBT) offers a competitive result although this model was sensitive to hyperparameters when compared with our RLF models. 

\paragraph{Consistency of Performance}
We also note the performance inconsistencies between competing baselines across datasets (\textit{e.g.} Adaboost significantly underperforms on the \textit{Chars74k} dataset, whilst offering one of the more competitive results on the \textit{MNIST} dataset). This again highlights the sensitivity of these competing models compared to RLFs which remain relatively robust.


\begin{table*}[t]
\begin{center}
\setlength\tabcolsep{0.40cm}
\resizebox{\columnwidth}{!}{%
\begin{tabular}{l c c c c c}
\specialrule{.2em}{.1em}{.1em}
&\thead{\textit{G50c}} &\thead{\textit{Letter}} &\thead{\textit{USPS}} &\thead{\textit{MNIST}} &\thead{Chars74k} \\
\hline
Adaboost & 36.56$\pm$1.30 & 3.98$\pm$0.11 & 5.89$\pm$0.22 & 3.20$\pm$0.08 & 24.99$\pm$0.23 \\
Random Forests & 18.91$\pm$1.33 & 5.16$\pm$0.14 & 5.86$\pm$0.17 & 3.18$\pm$0.06 & 21.14$\pm$0.15 \\
Extremely Randomized Forests & 18.91$\pm$1.33 & 4.77$\pm$0.12 & 5.67$\pm$0.18 & 3.36$\pm$0.09 & 21.39$\pm$0.18 \\
Gradient Boosted Trees & 18.90$\pm$1.27 & 4.70$\pm$0.18 & 5.93$\pm$0.27 & 3.15$\pm$0.05 & 17.59$\pm$0.29 \\
XGBoost-Large & 26.40$\pm$1.22 & 3.95$\pm$0.14 & 7.22$\pm$0.32 & 2.98$\pm$0.06 & 18.32$\pm$0.23 \\
Alternating Decision Forests & 18.71$\pm$1.27 & 3.52$\pm$0.12 & 5.59$\pm$0.16 & 2.71$\pm$0.10 & 16.67$\pm$0.21 \\
Residual Likelihood Forests (Ours) & \textbf{17.75$\pm$1.20} & \textbf{2.59$\pm$0.06} & \textbf{4.89$\pm$0.09} & \textbf{1.81$\pm$0.03} & \textbf{16.33$\pm$0.25} \\
\hline
\end{tabular}
}
\vspace{1.0em}
\caption{Overall performance of our method when compared with its main competitors on 5 datasets. The best performing methods are bolded. We train our RLF model using the same settings as described by~\cite{schulter2013alternating} for each of the 5 datasets shown.}
\label{tab:overall_comparison}
\end{center}
\end{table*}

\subsection{Overall Results}
In Table~\ref{tab:overall_comparison}, we list the overall empirical results which compare our RLF method against various ensemble baselines. We can see that RLF outperforms all other baselines across all 5 datasets. Each of the methods uses an ensemble of 100 trees of 15 tree depth, except for Gradient Boosted Trees, where we found that changing the default settings in the XGBoost~\cite{chen2016xgboost} implementation from the default settings of 100 trees at a tree depth of 3 resulted in lower performance. For completeness, we list the results of a GBT model with 100 trees and 15 depth (XGBoost-Large) as a comparison.

\begin{table*}[t]
\begin{center}
\setlength\tabcolsep{0.25cm}
\resizebox{\columnwidth}{!}{%
\begin{tabular}{l c c c c c c c c}
\specialrule{.2em}{.1em}{.1em}
& \multicolumn{4}{c}{Performance Error (\%)} & \multicolumn{4}{c}{Compression Ratio} \\
\cmidrule(l){2-5} \cmidrule(l){6-9}
Dataset & \thead{Refined-A} &\thead{Refined-E} &\thead{RLF-D} &\thead{RLF-S} &\thead{Refined-A} &\thead{Refined-E} &\thead{RLF-D} &\thead{RLF-S} \\
\hline
\textit{USPS} & 5.10$\pm$0.10 & 5.69$\pm$0.15 & \textbf{5.01$\pm$0.04} & \cellcolor{blue!25} 5.46$\pm$0.16 & 2.86 & 15.14 & \textbf{5.58} & \cellcolor{blue!25} 22.32 \\
\textit{Letter} & 2.98$\pm$0.15 & 4.33$\pm$0.08 & \textbf{2.68$\pm$0.06} & \cellcolor{blue!25} 4.30$\pm$0.22 & 2.33 & 30.32 & \textbf{3.49} & \cellcolor{blue!25} 446.42 \\
\textit{MNIST} & 2.05$\pm$0.02 & 2.95$\pm$0.03 & \textbf{1.81$\pm$0.03} & \cellcolor{blue!25} 2.41$\pm$0.05 & 6.29 & 76.92 & \textbf{6.98} & \cellcolor{blue!25} 111.61 \\
\textit{Chars74k} & \textbf{15.40$\pm$0.1} & \cellcolor{blue!25} 18.00$\pm$0.09 & 16.33$\pm$0.25 & 18.51$\pm$0.17 & 1.70 & 37.04 & \textbf{5.75} & \cellcolor{blue!25} 368.30 \\
\hline
\end{tabular}
}
\vspace{1.0em}
\caption{Residual Likelihood Forests compared with~\cite{ren2015global} for performance/model compression trade off on 4 datasets. For a fair comparison, the best performing method for each dataset comparing across Refined-A and RLF-D is \textbf{bolded}, and the best performing method comparing across Refined-E and RLF-S is \colorbox{blue!25}{shaded in blue}. We note that when compared to the best performing accurate model of~\cite{ren2015global}, our method (RLF-D) achieves higher performance on 3 out of the 4 datasets and achieves competitive results on the \textit{Chars74k} dataset, using a much smaller fraction of the model capacity of~\cite{ren2015global}. When compared to the economic model of~\cite{ren2015global}, our method is able to achieve better than or competitive performance whilst further reducing model capacity by more than an order of magnitude.} 
\label{tab:refined_forest_compare}
\end{center}
\end{table*}

\subsection{Comparison with Global Refined Forests}
A recent work that tries to infuse global information into the RF approach is the method of~\cite{ren2015global}, where global leaf refining along with leaf pruning enables mutual information to be accounted for as well as reducing model size. However, this approach still requires the training of the full RFs before iterative refinement and pruning can occur as global information is not injected until after the initial forest construction stage has completed. We compare the accurate refined model (Refined-A) and economic refined model (Refined-E) from~\cite{ren2015global} with a shallow and deep variant of RLF (RLF-S and RLF-D respectively as shown in Table~\ref{tab:max_tree_depths}). 
\begin{table}[h]
\begin{center}
\resizebox{0.35\columnwidth}{!}{%
\begin{tabular}{l c c c c}
\specialrule{.2em}{.1em}{.1em}
& \multicolumn{3}{c}{Maximum Tree Depth} \\
\cmidrule(l){2-5}
Dataset &\thead{\textit{USPS}} &\thead{\textit{Letter}} &\thead{\textit{MNIST}} &\thead{\textit{Chars74k}} \\
\hline
\textit{RF/ADF} & 10 & 15 & 15 & 25 \\
\textit{RLF-D} & 7 & 13 & 12 & 15 \\
\textit{RLF-S} & 5 & 6 & 7 & 10 \\
\hline
\end{tabular}
}
\vspace{1.0em}
\caption{Comparison of maximum tree depths.} 
\label{tab:max_tree_depths}
\end{center}
\end{table}


In Table~\ref{tab:refined_forest_compare}, we compare on model performance in classification error, as well as model compactness using the compression ratio defined in~\cite{ren2015global} between our approach and the approach in~\cite{ren2015global}. Similar to~\cite{ren2015global}, the compression ratio is defined relative to model capacity of the original RF~\cite{breiman2001random} and ADF~\cite{schulter2013alternating} which have similar model sizes. In terms of classification error, our deep model (RLF-D) handily outperforms the accurate version of the refined model in~\cite{ren2015global} on 3 out of the 4 datasets compared against. It is worth noting that even on the \textit{Chars74k} dataset, we achieve competitive results, despite limiting our maximum tree depth to 15 (compared to the 25 depth trees used in~\cite{ren2015global}). We note that our model demonstrates a drastic decrease in model capacity whilst only giving a minor trade off in accuracy. These results further highlight the efficiency of our RLF method in constructing compact models which offer strong classification performance.

\section{Conclusion}
\label{sec:conclusion}
We have proposed Residual Likelihood Forests which offers a new approach for optimizing the combining of weak learners in an ensemble learning framework. Our method shows that weak learners in an ensemble can be constructed to optimize a global loss in a complementary manner through the generation of residual likelihoods instead of probability distributions in the base tree weak learner. Empirically, we show that this allows for construction of much shallower and more compact models whilst yielding higher classification performance. 

\paragraph{Acknowledgments}
The authors would like to thank the anonymous reviewers for their useful and constructive comments. This work was supported by the Australian Research Council Centre of Excellence for Robotic Vision (project number CE1401000016).

\bibliography{egbib}
\end{document}